\documentclass[letterpaper]{article} 
\usepackage{arxiv}  
\usepackage{times}  
\usepackage{helvet}  
\usepackage{courier}  
\usepackage[hyphens]{url}  
\usepackage{graphicx} 
\usepackage{natbib}  
\usepackage{caption}
\usepackage{algorithm}
\usepackage{algorithmic}
\usepackage{newfloat}
\usepackage{listings}
\usepackage{amsmath}
\usepackage{amsfonts}
\usepackage{amsthm}
\usepackage{amssymb}
\usepackage{bbm}
\usepackage{bm}
\usepackage{multirow} 
\usepackage{booktabs}
\usepackage{color}
\usepackage{subfigure}
\usepackage{appendix}
\usepackage[colorlinks]{hyperref}
\usepackage{lipsum}
\newcommand\blfootnote[1]{
\begingroup
\renewcommand\thefootnote{}\footnote{#1}
\addtocounter{footnote}{-1}
\endgroup
}

\DeclareCaptionStyle{ruled}{labelfont=normalfont,labelsep=colon,strut=off}
\floatstyle{ruled}
\newfloat{listing}{tb}{lst}{}
\floatname{listing}{Listing}

\newcommand{\methodname}{ISL}

\title{Intra-Inter Subject Self-supervised Learning for Multivariate Cardiac Signals}
\author{
    Xiang Lan\textsuperscript{\rm 1},
    Dianwen Ng\textsuperscript{\rm 3},
    Shenda Hong\textsuperscript{\rm 4, \rm 5}\textsuperscript{$*$},
    Mengling Feng\textsuperscript{\rm 1, \rm 2}\textsuperscript{$*$}
    \blfootnote{Corresponding Authors}
}
\affiliations{
    \textsuperscript{\rm 1}Saw Swee Hock School of Public Health, National University of Singapore, Singapore\\
    \textsuperscript{\rm 2}Institute of Data Science, National University of Singapore, Singapore\\
    \textsuperscript{\rm 3}School of Computer Science and Engineering, Nanyang Technological University, Singapore\\
    \textsuperscript{\rm 4}National Institute of Health Data Science, Peking University, Beijing, China\\
    \textsuperscript{\rm 5}Institute of Medical Technology, Health Science Center of Peking University, Beijing, China\\
    \{ephlanx, ephfm\}@nus.edu.sg, 
    dianwen001@e.ntu.edu.sg, 
    hongshenda@pku.edu.cn
}
 
\usepackage{bibentry}
\begin{document}

\maketitle

\begin{abstract}
Learning information-rich and generalizable representations effectively from unlabeled multivariate cardiac signals to identify abnormal heart rhythms (cardiac arrhythmias) is valuable in real-world clinical settings but often challenging due to its complex temporal dynamics. Cardiac arrhythmias can vary significantly in temporal patterns even for the same patient ($i.e.$, intra subject difference). Meanwhile, the same type of cardiac arrhythmia can show different temporal patterns among different patients due to different cardiac structures ($i.e.$, inter subject difference). In this paper, we address the challenges by proposing an \textbf{I}ntra-inter \textbf{S}ubject self-supervised \textbf{L}earning (ISL) model that is customized for multivariate cardiac signals. Our proposed ISL model integrates medical knowledge into self-supervision to effectively learn from intra-inter subject differences. In intra subject self-supervision, ISL model first extracts heartbeat-level features from each subject using a channel-wise attentional CNN-RNN encoder. Then a stationarity test module is employed to capture the temporal dependencies between heartbeats. In inter subject self-supervision, we design a set of data augmentations according to the clinical characteristics of cardiac signals and perform contrastive learning among subjects to learn distinctive representations for various types of patients. Extensive experiments on three real-world datasets were conducted. In a semi-supervised transfer learning scenario, our pre-trained ISL model leads about 10\% improvement over supervised training when only 1\% labeled data is available, suggesting strong generalizability and robustness of the model.

\end{abstract}

\section{Introduction}

\begin{figure}[t]
    \centering
    \includegraphics[width=0.475\textwidth]{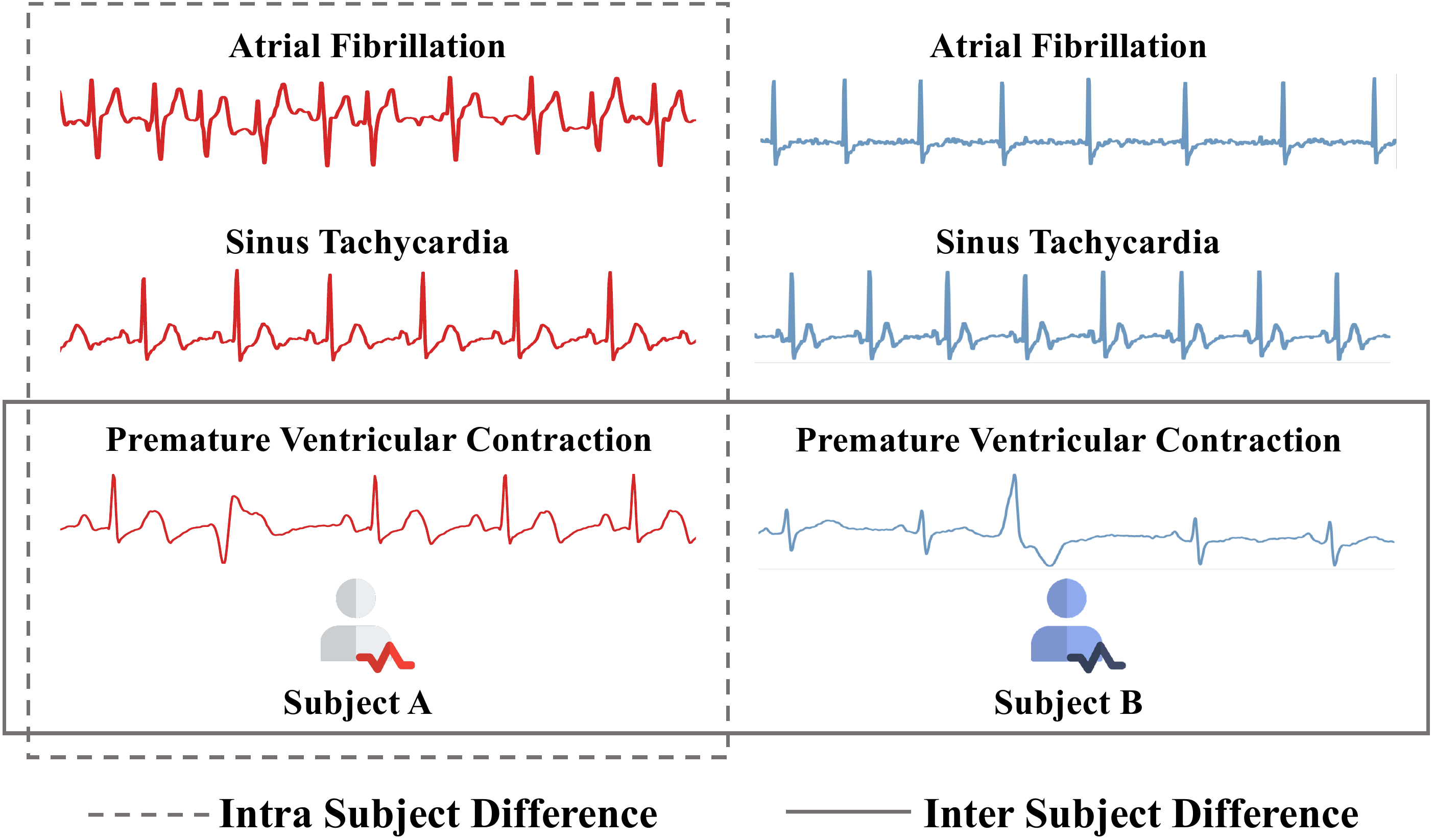}
    \caption{Example of the complex temporal dynamics of cardiac arrhythmias.
    The three different cardiac arrhythmias vary considerably in temporal patterns even if they all come from subject A ($i.e.$, intra subject difference). Also, because of different cardiac structures, the temporal features of Premature Ventricular Contraction between subject A and subject B are different ($i.e.$, inter subject difference).}
    \label{fig:dynamics.}
\end{figure}

Cardiovascular disease (CVD) is one of the primary causes of death globally. It has been reported with an estimation of 17.9 million deaths in 2019, representing 32\% of the entire global deaths \cite{who2019}. In clinical practice, 12-lead electrocardiography (ECG) is widely adopted to screen overall heart conditions \cite{doi:10.1161/CIRCULATIONAHA.106.180200}. Abnormal heart rhythms or heartbeats (cardiac arrhythmias) present in ECG may indicate anomaly cardiac functions that could result in severe CVD \cite{golany2019pgans}. Therefore, early and accurate diagnosis of ECG plays an important role in preventing severe CVD and can improve treatment outcomes \cite{169073}. Deep learning has been applied to improve the timeliness and accuracy of diagnosis of 12-lead ECG \cite{hong2020opportunities, mina, Zhu_2021}. However, training deep learning models in a supervised manner often requires a large volume of high-quality labels to achieve strong generalization performance. Despite the daily collection of thousands of ECG data by medical institutions, these data are mostly unlabeled. Moreover, it requires intensive labeling on the 12-lead ECG before being able to train a decent model. The work is taxing and requires huge efforts from domain experts, which is costly and not feasible in the real-world clinical setting. 

Self-supervised learning is a rising field and provides a promising solution to process unlabeled ECG data, whose main idea is to leverage data’s innate co-occurrence relationships as the self-supervision to mine useful representations \cite{Liu_2021}. Nevertheless, due to the complex temporal dynamics, learning information-rich and generalizable representations from unlabeled multivariate cardiac signals to identify cardiac arrhythmias remains to be a challenge. As illustrated in Figure \ref{fig:dynamics.}, we can observe two categories of differences: 

\begin{itemize}
\item \textbf{Intra Subject}: 
Different types of cardiac arrhythmias lead to significantly different temporal patterns in the cardia signals even for the same patient. For example, for Subject A in Figure \ref{fig:dynamics.}, Sinus Tachycardia is associated to a pattern of a faster heart beats; Atrial Fibrillation then leads to a fast and at the same time irregular pattern; Premature Ventricular Contraction then results in abrupt changes to sinus rhythm causing structural changes in the cardic signals.

\item \textbf{Inter Subject}: 
The same type of cardiac arrhythmia can show different temporal patterns within different patients because of variations in everyone's cardiac structures. For example, in Figure \ref{fig:dynamics.}, Subject A has higher QRS voltages ($i.e.$, the narrow spikes in the signal) and larger fluctuation of T waves ($i.e.$, bumps after QRS) than Subject B, even if they are experiencing same cardiac arrhythmias ($e.g.$, Premature Ventricular Contraction). These differences could be due to Subject A has stronger cardiac muscles that produce higher electrical impulses.

\end{itemize}

In the literature, most of the previous methods were proposed for image data \cite{chen2020simple, Chen_2021_CVPR}, and not much focus  has been put to the time series data. Thus neither the pretext tasks they used ($e.g.$, predicting image rotation angle) nor data augmentations they applied ($e.g.$, cropping the image into small patches) are suitable for time series such as cardiac signals. 

To address these gaps, we propose an \textbf{I}ntra-inter \textbf{S}ubject self-supervised \textbf{L}earning (ISL) model that is customized for multivariate cardiac signals. Our ISL model is an end-to-end model that integrates two different self-supervision procedures: the intra subject self-supervision and the inter subject self-supervision. Both procedures incorporate medical domain knowledge. In \textbf{intra subject self-supervision}, based on the fact that our heart rhythm consists of heartbeats, we segment the cardiac signal of each subject into multiple equal length frames ($i.e.$, heartbeat-level time windows). Meanwhile, inspired by the experience of cardiologist in cardiac arrhythmia diagnosis, we design a CNN-RNN encoder with channel-wise attention to extract features from each frame ($i.e.$, heartbeat-level features). After that, we train the encoder to maximize the similarity between similar frames of each subject by leveraging a stationarity test module. In \textbf{inter subject self-supervision}, we first design a set of data augmentations according to the clinical characteristics of cardiac signals. Then we fuse heartbeat-level features extracted from the encoder to obtain subject's representations. Lastly, we perform subject-wise contrastive learning to learn distinctive representations.

The main contributions of this work are summarized in below.
\begin{itemize}

\item We present ISL, a novel self-supervision model for learning information-rich and generalizable representations from unlabeled multivariate cardiac signals. 

\item We are the first work that integrates medical knowledge into self-supervision to boost the performance of cardiac arrhythmias diagnosis, which has great value in real-world applications.

\item We conducted extensive experiments on three public datasets to enable reproducibility. Experimental results demonstrate ISL outperforms current state-of-the-art methods in the downstream task under various scenarios.

\end{itemize}

\begin{figure*}[t]
    \centering
    \includegraphics[width=\textwidth]{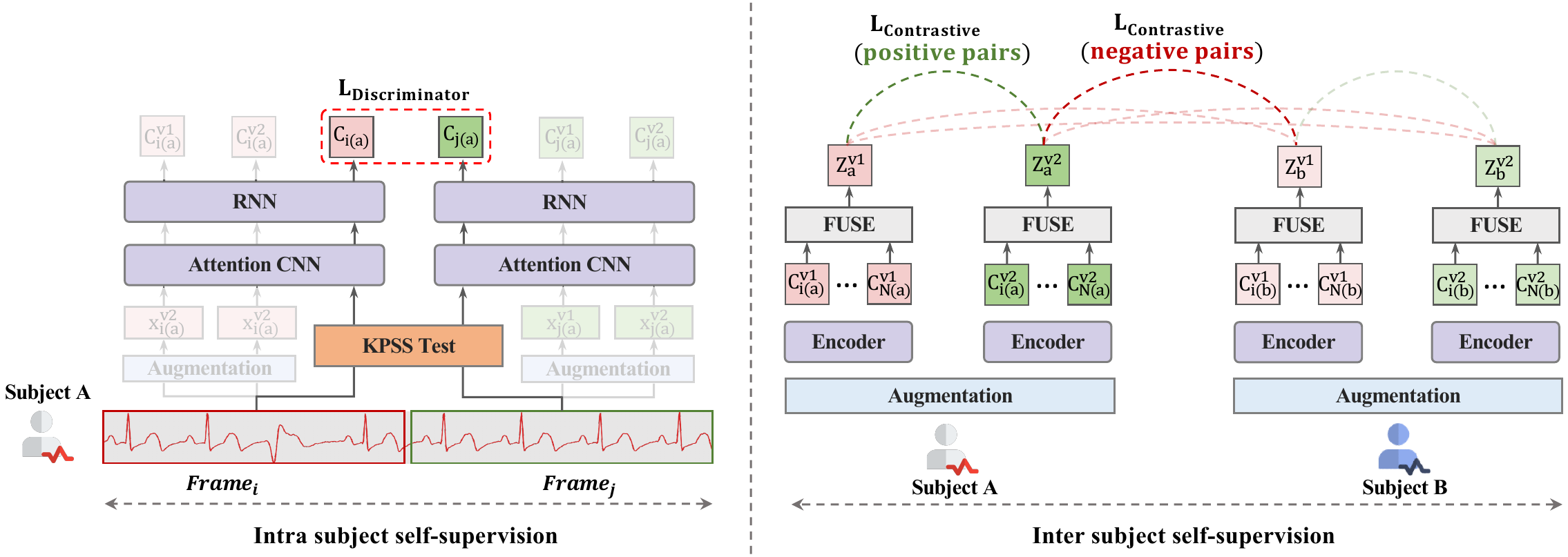}
    \caption{Overview of the ISL model which integrates two self-supervised learning procedures. \textbf{Intra subject self-supervision} aims to model the temporal dependencies within a cardiac signal. Take subject A as an example. The cardiac signal is first divided into $N$ equal length frames ($i.e.$, heartbeat-level time windows). Then, ISL uses the KPSS test to determine the stationarity of neighboring frames. The abrupt change occurs in $frame_i$ leading to a non-stationary time series. As such, $frame_i$ and $frame_j$ are treated as dissimilar frames. Meanwhile, ISL utilizes a channel-wise attentional CNN-RNN encoder to extract heartbeat-level features from each frame ($e.g.$, $\boldsymbol{c}_{i(a)}$ and $\boldsymbol{c}_{j(a)}$). These features are then fed into a discriminator to predict the probability of frames being similar.
    \textbf{Inter subject self-supervision} aims to learn distinctive representations among subjects. ISL first creates two views for each cardiac signal using specially designed data augmentations. After that, the heartbeat-level features extracted from the encoder are fused to be subjects' representations. Lastly, ISL performs contrastive learning among subjects ($e.g.$, Subject A and Subject B) to learn distinctive representations. Representations from the same subject are treated as positive pairs ($e.g.$, $\boldsymbol{z}_{a}^{v1}$ and $\boldsymbol{z}_{a}^{v2}$), while representations from different subjects are negative pairs ($e.g.$, $\boldsymbol{z}_{a}^{v2}$ and $\boldsymbol{z}_{b}^{v1}$). Finally, ISL jointly minimizes the discriminator loss and contrastive loss.}
    \label{fig:ISL}
\end{figure*}

\section{Methods}

\subsection{Overview}

In this section, we describe our approach in details. Figure \ref{fig:ISL} provides an overview of our proposed ISL model. We will elaborate on our methods of intra subject self-supervision and inter subject self-supervision in Section \ref{intra} and Section \ref{inter}, respectively. 

We represent the input multivariate cardiac signal as $\boldsymbol{X} \in R^{H \times L}$, where $H$ is the number of channels ($i.e.$, the number of leads in multi-lead ECG) and $L$ is the length of the signal. In our experiments, $H$$=$12 and $L$$=$5,000 ($i.e.$, 10 seconds duration of heartbeats). We divide each signal into $N$ equal length and non-overlapping frames, the $i-$th frame represented as $\boldsymbol{x}_i \in R^{H \times l}$. We set $N$$=$10 and $l$$=$500, such that each frame is 1-second duration, which contains about one to two heartbeats. 

Our goal is to train an encoder $F_{enc}(\theta_e)$ that projects $\boldsymbol{x}_i$ to a latent representation (Eq. \ref{frame_embed}) and finally obtain the full representation (Eq. \ref{subject_embed}) of a subject.

\begin{equation}
    \label{frame_embed}
    \boldsymbol{c}_i=F_{enc}(\boldsymbol{x}_i | \theta_e)
\end{equation} 
\begin{equation}
    \label{subject_embed}
    \boldsymbol{z_{\textbf{x}}}=\operatorname{Fuse}(\{\boldsymbol{c}_{1}, \boldsymbol{c}_{2},\hdots,\boldsymbol{c}_{N}\})
\end{equation} 
\subsection{Intra Subject Self-supervision}

\label{intra}
As shown in Figure \ref{fig:dynamics.}, the temporal patterns of cardiac arrhythmias can vary significantly even from the same subject. To capture such intra subject differences, we first design a channel-wise attentional CNN-RNN encoder to extract information-rich cross-channel features of the frames ($i.e.$, heartbeat-level features) from each subject (Eq. \ref{frame_embed}). Then we model the temporal dependencies between frames by utilizing a stationarity test module.

\subsubsection{Multivariate Cardiac Signal Encoding.}
Cardiologists may only look at specific channels for possible abnormal cardiac functions in clinical practice, not all 12 leads. For example, Posterior Wall Myocardial Infarction (MI) is only presented in four chest leads of a 12-lead ECG. This suggests that different cardiac arrhythmias should have different importance weighting for each channel. Therefore, we use the 1-dimensional convolution with channel-wise attention \cite{hu2018squeeze} for the encoder of our model to extract information-rich cross-channel features. Formally, given an input frame $\boldsymbol{x}$, it is first fed into a 1-dimensional convolutional layer and then compacted using Global Average Pooling (GAP). The latent features after GAP can be expressed as 
\begin{equation}
\label{eq3}
\boldsymbol{u} = \operatorname{GAP}( \operatorname{Conv1d}(\boldsymbol{x}))
\end{equation}
Thereafter, the channel weights $\boldsymbol{s}$ are calculated by 
\begin{equation}
\label{eq4}
\boldsymbol{s}=\operatorname{Sigmoid}(W_2 \cdot \operatorname{ReLU}(W_1 \cdot \boldsymbol{u}))
\end{equation}
The $h$-th channel of the output $\hat{\boldsymbol{x}}\in R^{H \times \frac{L}{r}}$ is re-weighted to 
\begin{equation}
\label{eq5}
\hat{\boldsymbol{x}}_h = \boldsymbol{s}_h \cdot \operatorname{Conv1d}(\boldsymbol{x}_h)
\end{equation}
Where $r$ is the reduction rate, $W_1$ and $W_2$ is the learnable parameters. To better capture the context of a frame, we add a two-layer Recurrent Neural Networks (RNN) on top of the convolutional layers. The final representation of an input frame is formulated as 
\begin{equation}
\label{eq6}
\boldsymbol{c}=\operatorname{RNN}(\hat{\boldsymbol{x}})
\end{equation}

\subsubsection{Stationarity Test Module.}
Next, to capture some cardiac arrhythmias that can cause abrupt changes to the heart rhythm, the intra-subject temporal dependencies between frames must be considered. For example, in Figure \ref{fig:ISL}, an abrupt change occurs in $frame_i$ while $frame_j$ remains normal. Even though these two frames are from the same subject, they cannot be regarded as a positive pair since they are not semantically similar. Therefore, the representations between dissimilar frames should be discriminated.

For this purpose, we first assume that abrupt changes in the cardiac signal can lead to a non-stationary time series. Then, we apply the Kwiatkowski–Phillips–Schmidt–Shin (KPSS) test \cite{kwiatkowski1992testing} to every neighboring frames to identify the occurrence of abrupt changes within the signal. The KPSS test is a statistical method for checking the stationarity of a time series using a null hypothesis that an observable time series is stationary. 

In our approach, given a pair of neighboring frames $\boldsymbol{x}_i$ and $\boldsymbol{x}_{i+1}$, we calculate the $p$-value for consecutive time series $(\boldsymbol{x}_{i}, \boldsymbol{x}_{i+1})$ using KPSS test. If the $p$-value from the test is below the threshold of 0.05, it means that the null hypothesis is rejected and suggests that the time series is non-stationary. We then use the test result as a pseudo label for each neighboring pair, and we train a discriminator $D(\theta_{d})$ to predict the stationarity of $(\boldsymbol{x}_{i}, \boldsymbol{x}_{i+1})$ using the concatenated representation $(\boldsymbol{c}_{i}, \boldsymbol{c}_{i+1})$. The learning objective of the discriminator is to minimize the loss defined in Eq. \ref{BCE}.
\begin{equation}
\label{BCE}
\mathcal{L}_{d} = -y \cdot \log \left(\hat{y}\right)-\left(1-y\right) \cdot \log \left(1-\hat{y}\right)
\end{equation}
Where $y$ and $\theta_d$ is the pseudo label from the stationarity test and discriminator's parameters, respectively. $y$ $=$ 1 if ($\boldsymbol{x}_i$, $\boldsymbol{x}_{i+1}$) is stationary time series, $y$ $=$ 0 otherwise. $\hat{y}$ $=$ $D((\boldsymbol{c}_{i},\boldsymbol{c}_{i+1})|\theta_d)$ is the estimated probability from the discriminator for a positive prediction. By doing so, the encoder is encouraged to generate distinguishable representations between similar and dissimilar frames, thus capture the intra subject temporal dependencies.

\subsection{Inter Subject Self-supervision}
\label{inter}
Another challenge is that cardiac arrhythmias can show different temporal patterns among patients due to the different cardiac structures or functions, even if the cardiac arrhythmias are coming from the same category. To address this challenge, we first fuse the heartbeat-level features (Eq. \ref{frame_embed}) to obtain a representation (Eq. \ref{subject_embed}) containing complete information of the cardiac signal ($i.e.$, heart rhythm-level feature). Then we perform contrastive learning among patients to effectively learn distinctive representations from inter subject differences.

\subsubsection{Multivariate Cardiac Signal Augmentation.}

An important component in contrastive learning is data augmentation that provides different views of data. The augmented data should preserve the semantic meaning of the raw data while providing additional information that are targeted to maximize the agreement. However, most prevalent data augmentations are designed for image data that may not be suitable for cardiac signals. Therefore, in addition to regular transformations such as flipping the signal or reversing the magnitude of the signal, we apply explicitly three additional data augmentations for multivariate cardiac signals. (See Appendix for more details of each augmentation.) 
\begin{itemize}
    \item \textbf{Baseline filtering}: We apply Daubechies 5 (db5) wavelet with a decomposition level of 5 to the signal. Then we obtain the signal baseline by reconstructing the signal using the approximation coefficients array from the fifth level decomposition. We use the signal baseline to provide a view that contains morphological information.
    
    \item \textbf{Bandpass filtering}: We apply Finite Impulse Response (FIR) bandpass filter \cite{signalsystem} to decompose the signal into low, middle, and high-frequency bands. The low-frequency band (0.001-0.5 Hz) preserves the shape of the signal, while the high-frequency band ($>$50 Hz) is mostly noise. Therefore, we select the middle-frequency band (0.5-50 Hz) as the transformed signal to provide a view of denoised signal.
    
    \item \textbf{Channel-wise difference}: Inspired by \cite{mohsenvand2020contrastive}, we subtract each adjacent channel of the cardiac signal to obtain a new channel that represents the voltage difference between two channels. In this way, our model is encouraged to learn the relationships between channels.
    
\end{itemize}

\subsubsection{Contrastive Learning.}
Contrastive learning aims to maximize the agreements of different yet relevant views from the same subject since they share the same underlying semantics and are regarded as positive pairs. Views from different subjects are treated as negative pairs, and the similarity between them should be minimized. We first generate two differently augmented data $\boldsymbol{X}^{v1}=g(\boldsymbol{X})$ and $\boldsymbol{X}^{v2}=g(\boldsymbol{X})$ for data $\boldsymbol{X}$, where the augmentations $g$ are randomly selected from our augmentation set $G = \{g_1,g_2,\hdots,g_t\}$. $\boldsymbol{X}^{v1}$ and $\boldsymbol{X}^{v2}$ are then divided into $N$ equal length frames $\{\boldsymbol{x}^{v1}_1, \boldsymbol{x}^{v1}_2, \hdots, \boldsymbol{x}^{v1}_{N}\}$ and $\{\boldsymbol{x}^{v2}_1, \boldsymbol{x}^{v2}_2, \hdots, \boldsymbol{x}^{v2}_{N}\}$. After that, the frames are fed into the encoder to extract heartbeat-level features $\{\boldsymbol{c}^{v1}_1, \boldsymbol{c}^{v1}_2, \hdots, \boldsymbol{c}^{v1}_{N}\}$ and $\{\boldsymbol{c}^{v2}_1, \boldsymbol{c}^{v2}_2, \hdots, \boldsymbol{c}^{v2}_{N}\}$ using Eq. \ref{eq3}, Eq. \ref{eq4}, Eq. \ref{eq5}, Eq. \ref{eq6}. Once we obtain the heartbeat-level features, we can fuse them to obtain the heart rhythm-level feature, which contains complete information to represent the subject. In our implementation, we simply use the aggregation of heartbeat-level features as the subject's representation, which can be denoted as $\boldsymbol{z}^{v1}_{\textbf{x}}=\sum_{i=1}^{N}\boldsymbol{c}_{i}^{v1}$ and $\boldsymbol{z}^{v2}_{\textbf{x}}=\sum_{i=1}^{N}\boldsymbol{c}_{i}^{v2}$. For a minibatch with $B$ subjects, representations from the same subject are positive pairs. We treat the other $2B-1$ representations from different subjects as negative pairs. The learning objective is to minimize the contrastive loss as described in Eq. \ref{NTXENT}.
\begin{equation}
\label{NTXENT}
\mathcal{L}_{c}=-\sum_{m=1}^{B} \log \frac{\exp \left(\operatorname{sim}\left(\boldsymbol{z}_{m}^{v1}, \boldsymbol{z}_{m}^{v2}\right) / \tau\right)}{\sum_{k=1}^{2 B} \mathbbm{1}_{[k \neq m]} \exp \left(\operatorname{sim}\left(\boldsymbol{z}_{m}^{v*}, \boldsymbol{z}_{k}^{v*}\right) / \tau\right)}
\end{equation}
Where $\tau$ is the temperature parameter, $\mathbbm{1}$ is an indicator function evaluating to 1 iff $k \neq m$. $\operatorname{sim}(\boldsymbol{u}, \boldsymbol{v})= \frac{\boldsymbol{u}^{\top} \boldsymbol{v}}{\|\boldsymbol{u}\|\|\boldsymbol{v}\|}$ calculate the cosine similarity between representation $\boldsymbol{u}$ and $\boldsymbol{v}$.

Lastly, combined with intra-inter subject self-supervision, we jointly minimize the self-supervision loss defined in Eq. \ref{BCE} and Eq. \ref{NTXENT}. The end-to-end training procedure of ISL is summarized in Algorithm \ref{ISL_algo}.

\begin{algorithm}[t]
\caption{Self-supervised training procedure of ISL}
\label{ISL_algo}
\textbf{Input}: Pre-training dataset $\mathcal{P}=\{\boldsymbol{X}_1, \boldsymbol{X}_2, ..., \boldsymbol{X}_p\}$, augmentation set $G$, pretraining iterations $Max\_iter$, number of frames $N$.\\
\textbf{Parameter}: ISL encoder $F_{enc}(\theta_e)$, ISL discriminator $D(\theta_d)$.\\
\textbf{Output}: Well-trained $F_{enc}(\hat{\theta}_e)$.
\begin{algorithmic}[1] 
\STATE {\bfseries Initialize} $\theta_e$, $\theta_{d}$
	    \FOR{$iter=0$ to $Max\_iter$:}
	        \STATE Sample a mini-batch $B$ from $\mathcal{P}$
	        \FOR{ $\boldsymbol{X}$ in $B$:}
	            \STATE Random sample two augmentations: $(g_1, g_2)\in G$ 
	            \STATE Create two views: $\boldsymbol{X}^{v1}$, $\boldsymbol{X}^{v2}$ $\gets$ $g_1(\boldsymbol{X})$, $g_2(\boldsymbol{X})$
	            \STATE Divide signal into equal length frames:
	            \STATE $\{\boldsymbol{x}_{1}, \boldsymbol{x}_2, \hdots, \boldsymbol{x}_{N}\} $ $\gets$ $\boldsymbol{X}$\\
	            $\{\boldsymbol{x}^{v1}_1, \boldsymbol{x}^{v1}_2, \hdots, \boldsymbol{x}^{v1}_{N}\} $ $\gets$ $\boldsymbol{X}^{v1}$\\
	            $\{\boldsymbol{x}^{v2}_1, \boldsymbol{x}^{v2}_2, \hdots, \boldsymbol{x}^{v2}_{N}\} $ $\gets$ $\boldsymbol{X}^{v2}$
	            
	            \WHILE{$i \leq (N-1)$}
	            \STATE $p_{i} =$ KPSS($\boldsymbol{x}_{i}$, $\boldsymbol{x}_{i+1}$)
	            \STATE
                $y_{i}=\left\{\begin{array}{ll}1, & \text {if } p_{i} \geq 0.05 \\ 0, & \text {if } p_{i}<0.05\end{array}\right.$
                \STATE $\hat{y}_i = D(F_{enc}((\boldsymbol{x}_{i}$, $\boldsymbol{x}_{i+1}), \theta_e),\theta_d)$
                \STATE Minimize the loss $\mathcal{L}_{d}$ in Eq. \ref{BCE}
	            \ENDWHILE
            \STATE Get $\boldsymbol{c}^{v1}_{i}$,  $\boldsymbol{c}^{v2}_{i}$ by Eq. \ref{eq3}, Eq. \ref{eq4}, Eq. \ref{eq5}, Eq. \ref{eq6}.
	        \STATE Get representations of the subject: \\ $\boldsymbol{z}^{v1}_{\textbf{x}}=\sum_{i=1}^{N}\boldsymbol{c}_{i}^{v1}$, $\boldsymbol{z}^{v2}_{\textbf{x}}=\sum_{i=1}^{N}\boldsymbol{c}_{i}^{v2}$
            \ENDFOR
        \STATE Minimize the loss $\mathcal{L}_{c}$ in Eq. \ref{NTXENT}
	    \ENDFOR
	    \STATE \textbf{return} $\hat{\theta}_e$
	\end{algorithmic}  
\end{algorithm}

\section{Experiment Setup}

\begin{table}[t]
\footnotesize
\centering
\begin{tabular}{l | c c c c}
\toprule
Dataset&\multicolumn{1}{c}{Train}&\multicolumn{1}{c}{Validation}&\multicolumn{1}{c}{Test}&\multicolumn{1}{c}{Categories}\\
\midrule
Chapman         & 6,352  &  2,113  & 2,123 & 4 \\
CPSC            & 5,612  &  1,870  & 1,818 & 9 \\
PTB-XL          & 13,104 &  4,361  & 4,370 & 71\\
\bottomrule 
\end{tabular}
\caption{Number of samples and cardiac arrhythmia categories in each pre-processed dataset used in our experiments.}
\label{table:dataset}
\end{table}

\subsection{Datasets}
To benchmark the performance of our proposed ISL model and to ensure reproducibility of our results, we pick three of the largest publicly available real-world ECG datasets for cardiac arrhythmias classification. We split each dataset into 60\%, 20\%, 20\% in subject-wise for training, validation and testing. Table \ref{table:dataset} shows description of each pre-processed dataset. More details of data pre-processing are provided in Appendix.

\subsubsection{Chapman.}
Chapman \cite{zheng202012} contains 12-lead ECG recordings of 10,646 patients with a sampling rate of 500 Hz. The dataset includes 11 common cardiac arrhythmias. Each recording length is 10 seconds, and we grouped these cardiac arrhythmias into four categories for a fair comparison with \cite{kiyasseh2020clocs}.

\subsubsection{CPSC.}
CPSC \cite{liu2018open} contains 12-lead ECG recordings of 6,877 patients with a sampling rate of 500 Hz. The dataset covers nine types of cardiac arrhythmias. The recording length is from 6 seconds to 60 seconds. In our setting, we truncated each record to the same length of 10 seconds.

\subsubsection{PTB-XL.}
PTB-XL \cite{wagner2020ptb} contains 21,837 12-lead ECG recordings from 18,885 patients with a sampling rate of 500 Hz. The recording length is 10 seconds and covers total 71 different cardiac arrhythmias.

\subsection{Baselines}
We compare our method with the following baselines. 
(1) \textbf{Random Init.:} Training a logistic regression model using features extracted from randomly initialized ISL encoder.
(2) \textbf{Supervised:} Pre-training ISL in a supervised manner.
(3) \textbf{CPC}\cite{oord2018representation}.
(4) \textbf{BYOL}\cite{grill2020bootstrap}. 
(5) \textbf{SimCLR}\cite{chen2020simple}.
(6) \textbf{SSLECG}\cite{sarkar2020self}.
(7) \textbf{CLOCS}\cite{kiyasseh2020clocs}.

Since CPC, BYOL, and SimCLR are termed as general self-supervised learning frameworks, for a fair comparison, we investigated the performance of these frameworks using the ISL encoder and the basic encoder used in their paper and report their best performance. Furthermore, we applied the same data augmentation as ISL to BYOL and SimCLR since their original augmentations were designed only for image data. The target decay rate of BYOL is set to 0.996. The temperature parameter of SimCLR is set to 0.1. For the implementation of SSLECG and CLOCS, we referred to their published codes during execution. In our comparison, the embedding dimension of all methods are set to $E$=256.

\subsection{Downstream Evaluation Task}
Accurate diagnosis of cardiac arrhythmias is of great importance in clinical workflows. Therefore, the downstream task is a multi-label classification of cardiac arrhythmias. Formally, given the subject representation $\boldsymbol{z}_{\textbf{x}}$, we train a logistic regression model with parameters $\theta_{lr}$ to identify all cardiac arrhythmias presented in the cardiac signal $\boldsymbol{X}$ of a subject:  
\begin{equation}
     \operatorname{Predictions} = \operatorname{LogisticRegression}(\boldsymbol{z}_{\textbf{x}}|\theta_{lr})
\end{equation}

We comprehensively evaluate our ISL model under three scenarios. 
\subsubsection{Linear Evaluation of Representations.} We follow the standard linear evaluation scheme in \cite{bert,chen2020simple}, where the pre-trained encoder is frozen, and a logistic regression model is trained on the downstream task in a supervised manner.

\subsubsection{Transferability Evaluation of Representations.} This transfer learning scenario aims to evaluate the robustness and generalizability of the learned representations. In this setting, we pre-train the model on one dataset and perform supervised fine-tuning on the other two downstream datasets, respectively.

\subsubsection{Semi-supervised Learning Experiments.} To simulate the real-world situation that some medical institutions may have limited labeled data, we investigate the performance of our pre-trained model in a label-scarce scenario. In this setting, we pre-train our model on one dataset and fine-tune the model with different percentages of labeled data on another dataset.

\subsection{Implementation Details}
The model is optimized using Adam optimizer \cite{adam} with a learning rate of 3e-3 and weight decay of 4e-4. We use a hard-stop of 40 epochs and a batch size of 232 for both pre-training and downstream tasks, as the training loss does not further decrease. The experiments were conducted using PyTorch 1.8 \cite{paszke2019pytorch} on NVIDIA GeForce Tesla V100 GPU, we run each experiment 5 times with 5 different seeds, and we report the mean and standard deviation of the test Area Under the Receiver Operating Characteristic Curve (AUROC).

\section{Experimental Results}
\subsection{Linear Evaluation}

\begin{table}[t]
\footnotesize
\centering
\begin{tabular}{l c c c}
\toprule
Dataset&\multicolumn{1}{c}{Chapman}&\multicolumn{1}{c}{CPSC}&\multicolumn{1}{c}{PTB-XL}\\
\midrule
Supervised  &   0.990$\pm$0.001  & 0.888$\pm$0.012 & 0.781$\pm$0.010\\
Random Init. &   0.825$\pm$0.006  & 0.588$\pm$0.007 & 0.549$\pm$0.019\\
\midrule
CPC       &  0.844$\pm$0.019 & 0.711$\pm$0.030 & 0.628$\pm$0.027\\
BYOL      &  0.592$\pm$0.014 & 0.583$\pm$0.042 & 0.539$\pm$0.023\\
SimCLR    &  0.771$\pm$0.027 & 0.619$\pm$0.008 & 0.631$\pm$0.013\\
SSLECG    &  0.526$\pm$0.026 & 0.512$\pm$0.014 & 0.476$\pm$0.036\\
CLOCS  &  0.906$\pm$0.003 & 0.764$\pm$0.011 & 0.619$\pm$0.020\\
\midrule
\textbf{\methodname (w/o Inter)}  & 0.764$\pm$0.011 & 0.700$\pm$0.009 &  0.660$\pm$0.011\\
\textbf{\methodname (w/o Intra)} & 0.921$\pm$0.030 & 0.825$\pm$0.034 & 0.713$\pm$0.024 \\
\textbf{\methodname} & \textbf{0.965$\pm$0.008} & \textbf{0.854$\pm$0.012} & \textbf{0.722$\pm$0.012}\\
\bottomrule 
\end{tabular}
\caption{Test AUROC of the linear evaluation.}
\label{table:linear}
\end{table}

\begin{table}[t]
\footnotesize
\centering
\begin{tabular}{l c c c}
\toprule
Dataset&\multicolumn{1}{c}{Chapman}&\multicolumn{1}{c}{CPSC}&\multicolumn{1}{c}{PTB-XL}\\
\midrule
\textbf{\methodname (\textit{E}=32)}  & 0.908$\pm$0.012  & 0.715$\pm$0.017  & 0.583$\pm$0.021\\
\textbf{\methodname (\textit{E}=64)}  & 0.930$\pm$0.015  & 0.801$\pm$0.014 & 0.621$\pm$0.018\\
\textbf{\methodname (\textit{E}=128)} & 0.915$\pm$0.034  & 0.833$\pm$0.007  & 0.633$\pm$0.035\\
\bottomrule 
\end{tabular}
\caption{Test AUROC of the linear evaluation with different embedding dimensions.}
\label{table:linear_ablation}
\end{table}

We conduct a linear evaluation to evaluate the quality of representations learned by ISL. In Table \ref{table:linear}, ISL outperforms other state-of-the-art approaches on all three datasets, and the performance is comparable with supervised training. For example, ISL has a 9\% and 10.3\% improvement compared with CLOCS on CPSC and PTB-XL datasets, respectively. On the Chapman dataset, the performance gap between ISL and supervised training is only 2.5\%. Furthermore, in Table \ref{table:linear_ablation}, we show that our model performs consistently well even when we reduced the dimension of the representation. Our ISL still achieves a decent AUROC score when the representation's dimension decreased to 32. Lower embedding dimension means less computational consumption and faster pre-training. This property might be useful in real-world applications where some hospitals have limited computational power. Overall, the results of our linear evaluation imply that, in comparison to other state-of-the-art methods,  representations learned by ISL are richer in information.

\subsection{Transferability Evaluation}

\begin{table*}[t]
\footnotesize
\centering
\begin{tabular}{l | c c | c c | c c }
\toprule
\multirow{1}{*}{Pre-training Dataset}&\multicolumn{2}{c}{Chapman}&\multicolumn{2}{c}{CPSC}&\multicolumn{2}{c}{PTB-XL}\\
\midrule
Downstream Dataset & CPSC & PTB-XL & Chapman & PTB-XL & Chapman & CPSC\\
\midrule
Supervised         & 0.797 $\pm$ 0.013 & 0.723 $\pm$ 0.008 & 0.919 $\pm$ 0.009 & 0.757 $\pm$ 0.011 & 0.818 $\pm$ 0.050 &  0.777 $\pm$ 0.010\\
\midrule
CPC       & 0.868 $\pm$ 0.003  & 0.824 $\pm$ 0.009 & 0.887 $\pm$ 0.004 & \textbf{0.850 $\pm$ 0.010} & 0.874 $\pm$ 0.010 & 0.884 $\pm$ 0.003 \\

BYOL      & 0.865 $\pm$ 0.033 & 0.695 $\pm$ 0.093 & 0.958 $\pm$ 0.049 & 0.638 $\pm$ 0.044 & 0.885 $\pm$ 0.185 & 0.865 $\pm$ 0.064 \\

SimCLR    &  0.579 $\pm$ 0.024 & 0.568 $\pm$ 0.031 &  0.834 $\pm$ 0.071 & 0.643 $\pm$ 0.005 & 0.574 $\pm$ 0.016 & 0.551 $\pm$ 0.034 \\

SSLECG    &  0.922 $\pm$ 0.004 & 0.784 $\pm$ 0.019 & 0.977 $\pm$ 0.005 & 0.742 $\pm$ 0.052  & 0.977 $\pm$ 0.003  & 0.896 $\pm$ 0.023 \\

CLOCS    &  0.843 $\pm$ 0.006 & 0.740 $\pm$ 0.007 &  0.957 $\pm$ 0.004 & 0.741 $\pm$ 0.004 & 0.948 $\pm$ 0.006  & 0.775 $\pm$ 0.003 \\

\midrule
\textbf{\methodname (w/o Inter)}  & 0.895 $\pm$ 0.011 & 0.762 $\pm$ 0.022 & 0.987 $\pm$ 0.002 & 0.794 $\pm$ 0.009 & 0.989 $\pm$ 0.001 & 0.843 $\pm$ 0.106 \\
\textbf{\methodname (w/o Intra)}  & 0.914 $\pm$ 0.014 & 0.819 $\pm$ 0.024 & 0.989 $\pm$ 0.002 & 0.821 $\pm$ 0.017 & 0.989 $\pm$ 0.003 & 0.917 $\pm$ 0.011 \\
\textbf{\methodname} & \textbf{0.928 $\pm$ 0.003} & \textbf{0.831 $\pm$ 0.009} & \textbf{0.991 $\pm$ 0.001} & 0.830 $\pm$ 0.010 & \textbf{0.990 $\pm$ 0.002} &  \textbf{0.926 $\pm$ 0.003}\\
\bottomrule 
\end{tabular}
\caption{Test AUROC of the transferability evaluation.}
\label{table:transfer_evaluation}
\end{table*}

We evaluate the generalizability of representations learned by ISL in transfer learning scenarios. Table \ref{table:transfer_evaluation} shows the performance comparisons under six transfer learning settings. In general, ISL outperforms the other state-of-the-art approaches in five out of six settings. We find that ISL brings a 4\% and 5\% improvement on CPSC and PTB-XL, respectively, compared with supervised training in Table \ref{table:linear}. This suggests that the representations learned by ISL are robust and can generalize to other data sources to improve the performance further. We find that BYOL and SimCLR did not perform well in some of the transfer settings. We hypothesize that this may be due to the augmentations used during training. It is likely that BYOL and SimCLR adopts better to the set of augmentations built for images as compared to the ISL's augmentations techniques. We also find CPC performs consistently well compared with BYOL and SimCLR, which could be due to CPC taking intra subject temporal dependencies into account, while BYOL and SimCLR did not. However, CPC ignores the inter subject differences, thus the learned representations are not as generalizable as ISL.

\subsection{Semi-supervised Learning Experiments}
\begin{figure}[t]
    \centering
    \includegraphics[width=0.48\textwidth]{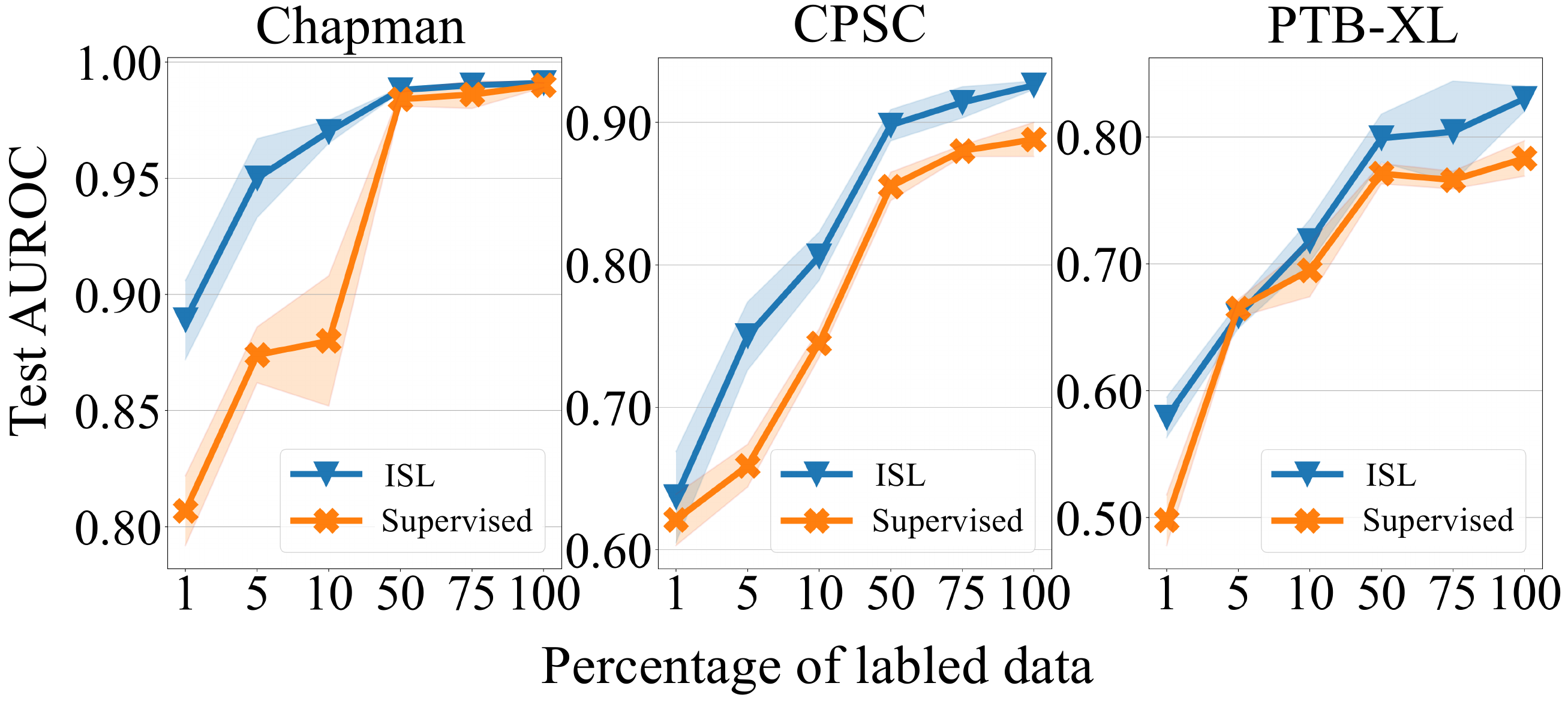}
    \caption{Test AUROC of semi-supervised learning experiments.}
    \label{fig:semi}
\end{figure}

To further study the effects of different percentages of labeled data, we conduct semi-supervised learning experiments. As shown in Figure \ref{fig:semi}, we find that ISL shows significant advantages in the label-scarce scenario. For example, the pre-trained ISL has about 10\% improvement on the Chapman dataset compared to supervised training when the percentage of labeled data is less than 50\%. We can observe that the pre-trained ISL model constantly performs better than supervised training under the semi-supervised learning scenario on three datasets.

\subsection{Ablation Analysis}
We study the contribution of the two self-supervision procedures in an ablation analysis as presented in Table \ref{table:linear} and Table \ref{table:transfer_evaluation}. 
\begin{itemize}
    \item  \textbf{w/o Inter}: In this setting, we remove the inter subject contrastive learning, and perform the intra subject self-supervision only. The learning objective is to minimize the loss defined in Eq. \ref{BCE}.
    
    \item  \textbf{w/o Intra}: In this setting, we remove the KPSS Test and the discriminator used in the intra subject self-supervision and perform inter subject contrastive learning only. The learning objective is to minimize the loss defined in Eq. \ref{NTXENT}.

\end{itemize}

The results reveal that, in both linear evaluation and transferability evaluation, inter subject self-supervision achieves a better performance than intra subject self-supervision, while intra subject self-supervision provides additional improvements to overall performance. Such results could be explained from two aspects. First, the intra subject self-supervision is based on a pretext task, which may limit the generalizability of the learned representations \cite{chen2020simple}. Second, when used solely with the intra subject self-supervision, the performance could be affected by the sample size of cardiac arrhythmias that cause abrupt changes of heartbeats. For example, the number of subjects that have Premature Ventricular Contraction is only 9\% of the total subjects in the CPSC dataset, resulting in imbalanced training examples for the discriminator. This issue could be alleviated by training on datasets with more non-stationary cardiac signal samples, such as the PTB-XL dataset. For example, in Table \ref{table:linear}, the performance gap between the two self-supervision procedures on PTB-XL dataset is much narrower than on the Chapman and CPSC dataset.

\section{Related works}
\subsection{Self-supervised Contrastive Learning}

Self-supervised contrastive learning has recently shown great promises and achieved state-of-the-art performance in many tasks. The motivation of self-supervised contrastive learning is to excavate shared information from different data transformations. For this purpose, multiple views or data augmentations are created for a sample. The learning objective is to maximize the mutual information between different augmentations of the same sample in the latent space. In view of this methodology, \cite{tian2020contrastive} propose contrastive multiview coding to learn invariant representations between different views of the same scene. \cite{He_2020_CVPR} present momentum contrast (MoCo) where the feature extractor is trained in a dictionary look-up manner. \cite{chen2020simple} present SimCLR that removed the memory bank. More recently, self-distillation based methods such as BYOL \cite{grill2020bootstrap} and DINO \cite{caron2021emerging} are proposed, where negative pairs are no longer necessary.

\subsection{Self-supervision for Physiological Signals}

Some works have studied self-supervision in time series data \cite{yue2021learning,mehari2021self,spathis2020learning,banville2019self,ma2019learning,franceschi2019unsupervised, hyvarinen2016unsupervised}. While a few works have recently explored the effectiveness of self-supervision for physiological signals. \cite{mohsenvand2020contrastive, eldele2021time} proposed self-supervised learning frameworks that tested with brain waves data. \cite{sarkar2020self} proposed a multi-task ECG representation learning framework for emotion recognition, where the model is pre-trained by a pretext task of predicting six handcrafted data transformations. \cite{oord2018representation} presented contrastive predictive coding (CPC) for speech recognition by predicting the near future state in a given utterance.

One work that is relevant to our method is TNC \cite{tonekaboni2021unsupervised}, a general unsupervised learning framework modeling the progression of temporal dynamics of time series. However, TNC pre-train the model by only predicting predefined neighboring relationships between time windows, thus having limited capacity in modeling complex cardiac signals. In comparison, ISL simultaneously models the intra subject temporal dependencies and inter subject differences so that the learned representations are more generalizable and robust. Moreover, TNC is mainly designed for univariate time series and is hard to generalize to 12-lead ECG data.

Another work relevant to our method is CLOCS \cite{kiyasseh2020clocs}, a contrastive learning model designed explicitly for multivariate cardiac signals. ISL differs CLOCS from the following aspects. First, CLOCS assumes that abrupt changes are unlikely to occur in heartbeats in few seconds. Thus all segments from the same cardiac signal share the same context in latent space. In other words, segments from the same subject are positive pairs. CLOCS also assumes that different channels of the same signal are positive pairs because they share the same underlying states. While as illustrated in Figure \ref{fig:dynamics.}, we did not make such assumptions for ISL since cardiac arrhythmias' patterns are multifarious. Instead, we consider all possible cases where abnormalities could occur suddenly or in specific ECG leads so that the learned representations can be more comprehensive. Furthermore, CLOCS directly learns subject's representations from the entire cardiac signal to perform contrastive learning. In contrast, ISL first learns heartbeat-level features and later fuses them to represent the subject. In this way, ISL is able to capture full-scale temporal dynamics of cardiac signals. 

\section{Conclusion}
In this paper, we propose a novel self-supervision model, ISL, for learning information-rich and generalizable representations from unlabeled multivariate cardiac signals to improve cardiac arrhythmias diagnosis. Our ISL model integrates intra subject self-supervision and inter subject self-supervision. The intra subject self-supervision procedure addresses the issue that temporal patterns differ considerably between cardiac arrhythmias, even from the same patient. The inter subject self-supervision procedure addresses the problem that the same type of cardiac arrhythmia shows different temporal patterns between patients due to different cardiac structures. Extensive experiments on three real-world datasets were conducted, the results over different evaluation scenarios show that the representation learned by ISL is information-rich and more generalizable than other state-of-the-art methods. Moreover, the results in label-scarce scenario suggest strong potential of ISL in real clinical applications.


\bibliography{ref}

\begin{thebibliography}{37}
\providecommand{\natexlab}[1]{#1}

\bibitem[{Artis, Mark, and Moody(1991)}]{169073}
Artis, S.; Mark, R.; and Moody, G. 1991.
\newblock Detection of atrial fibrillation using artificial neural networks.
\newblock In \emph{[1991] Proceedings Computers in Cardiology}, 173--176.

\bibitem[{Banville et~al.(2019)Banville, Albuquerque, Hyv{\"a}rinen, Moffat,
  Engemann, and Gramfort}]{banville2019self}
Banville, H.; Albuquerque, I.; Hyv{\"a}rinen, A.; Moffat, G.; Engemann, D.-A.;
  and Gramfort, A. 2019.
\newblock Self-supervised representation learning from electroencephalography
  signals.
\newblock In \emph{2019 IEEE 29th International Workshop on Machine Learning
  for Signal Processing (MLSP)}, 1--6. IEEE.

\bibitem[{Caron et~al.(2021)Caron, Touvron, Misra, J{\'e}gou, Mairal,
  Bojanowski, and Joulin}]{caron2021emerging}
Caron, M.; Touvron, H.; Misra, I.; J{\'e}gou, H.; Mairal, J.; Bojanowski, P.;
  and Joulin, A. 2021.
\newblock Emerging properties in self-supervised vision transformers.
\newblock \emph{arXiv preprint arXiv:2104.14294}.

\bibitem[{Chen et~al.(2020)Chen, Kornblith, Norouzi, and
  Hinton}]{chen2020simple}
Chen, T.; Kornblith, S.; Norouzi, M.; and Hinton, G. 2020.
\newblock A simple framework for contrastive learning of visual
  representations.
\newblock In \emph{International conference on machine learning}, 1597--1607.
  PMLR.

\bibitem[{Chen and He(2021)}]{Chen_2021_CVPR}
Chen, X.; and He, K. 2021.
\newblock Exploring Simple Siamese Representation Learning.
\newblock In \emph{Proceedings of the IEEE/CVF Conference on Computer Vision
  and Pattern Recognition (CVPR)}, 15750--15758.

\bibitem[{Devlin et~al.(2019)Devlin, Chang, Lee, and Toutanova}]{bert}
Devlin, J.; Chang, M.; Lee, K.; and Toutanova, K. 2019.
\newblock {BERT:} Pre-training of Deep Bidirectional Transformers for Language
  Understanding.
\newblock In Burstein, J.; Doran, C.; and Solorio, T., eds., \emph{Proceedings
  of the 2019 Conference of the North American Chapter of the Association for
  Computational Linguistics: Human Language Technologies, {NAACL-HLT} 2019,
  Minneapolis, MN, USA, June 2-7, 2019, Volume 1 (Long and Short Papers)},
  4171--4186. Association for Computational Linguistics.

\bibitem[{Eldele et~al.(2021)Eldele, Ragab, Chen, Wu, Kwoh, Li, and
  Guan}]{eldele2021time}
Eldele, E.; Ragab, M.; Chen, Z.; Wu, M.; Kwoh, C.~K.; Li, X.; and Guan, C.
  2021.
\newblock Time-Series Representation Learning via Temporal and Contextual
  Contrasting.
\newblock \emph{IJCAI}.

\bibitem[{Franceschi, Dieuleveut, and Jaggi(2019)}]{franceschi2019unsupervised}
Franceschi, J.-Y.; Dieuleveut, A.; and Jaggi, M. 2019.
\newblock Unsupervised Scalable Representation Learning for Multivariate Time
  Series.
\newblock \emph{Advances in Neural Information Processing Systems}, 32:
  4650--4661.

\bibitem[{Golany and Radinsky(2019)}]{golany2019pgans}
Golany, T.; and Radinsky, K. 2019.
\newblock Pgans: Personalized generative adversarial networks for ecg synthesis
  to improve patient-specific deep ecg classification.
\newblock In \emph{Proceedings of the AAAI Conference on Artificial
  Intelligence}, volume~33, 557--564.

\bibitem[{Grill et~al.(2020)Grill, Strub, Altch{\'e}, Tallec, Richemond,
  Buchatskaya, Doersch, Pires, Guo, Azar et~al.}]{grill2020bootstrap}
Grill, J.-B.; Strub, F.; Altch{\'e}, F.; Tallec, C.; Richemond, P.~H.;
  Buchatskaya, E.; Doersch, C.; Pires, B.~A.; Guo, Z.~D.; Azar, M.~G.; et~al.
  2020.
\newblock Bootstrap your own latent: A new approach to self-supervised
  learning.
\newblock \emph{arXiv preprint arXiv:2006.07733}.

\bibitem[{He et~al.(2020)He, Fan, Wu, Xie, and Girshick}]{He_2020_CVPR}
He, K.; Fan, H.; Wu, Y.; Xie, S.; and Girshick, R. 2020.
\newblock Momentum Contrast for Unsupervised Visual Representation Learning.
\newblock In \emph{Proceedings of the IEEE/CVF Conference on Computer Vision
  and Pattern Recognition (CVPR)}.

\bibitem[{Hong et~al.(2019)Hong, Xiao, Ma, Li, and Sun}]{mina}
Hong, S.; Xiao, C.; Ma, T.; Li, H.; and Sun, J. 2019.
\newblock MINA: Multilevel Knowledge-Guided Attention for Modeling
  Electrocardiography Signals.
\newblock In \emph{Proceedings of the Twenty-Eighth International Joint
  Conference on Artificial Intelligence, {IJCAI-19}}, 5888--5894. International
  Joint Conferences on Artificial Intelligence Organization.

\bibitem[{Hong et~al.(2020)Hong, Zhou, Shang, Xiao, and
  Sun}]{hong2020opportunities}
Hong, S.; Zhou, Y.; Shang, J.; Xiao, C.; and Sun, J. 2020.
\newblock Opportunities and challenges of deep learning methods for
  electrocardiogram data: A systematic review.
\newblock \emph{Computers in Biology and Medicine}, 103801.

\bibitem[{Hu, Shen, and Sun(2018)}]{hu2018squeeze}
Hu, J.; Shen, L.; and Sun, G. 2018.
\newblock Squeeze-and-excitation networks.
\newblock In \emph{Proceedings of the IEEE conference on computer vision and
  pattern recognition}, 7132--7141.

\bibitem[{Hyvarinen and Morioka(2016)}]{hyvarinen2016unsupervised}
Hyvarinen, A.; and Morioka, H. 2016.
\newblock Unsupervised feature extraction by time-contrastive learning and
  nonlinear ica.
\newblock \emph{Advances in Neural Information Processing Systems}, 29:
  3765--3773.

\bibitem[{Kingma and Ba(2015)}]{adam}
Kingma, D.~P.; and Ba, J. 2015.
\newblock Adam: {A} Method for Stochastic Optimization.
\newblock In Bengio, Y.; and LeCun, Y., eds., \emph{3rd International
  Conference on Learning Representations, {ICLR} 2015, San Diego, CA, USA, May
  7-9, 2015, Conference Track Proceedings}.

\bibitem[{Kiyasseh, Zhu, and Clifton(2021)}]{kiyasseh2020clocs}
Kiyasseh, D.; Zhu, T.; and Clifton, D.~A. 2021.
\newblock CLOCS: Contrastive Learning of Cardiac Signals Across Space, Time,
  and Patients.
\newblock In Meila, M.; and Zhang, T., eds., \emph{Proceedings of the 38th
  International Conference on Machine Learning}, volume 139 of
  \emph{Proceedings of Machine Learning Research}, 5606--5615. PMLR.

\bibitem[{Kligfield et~al.(2007)Kligfield, Gettes, Bailey, Childers, Deal,
  Hancock, van Herpen, Kors, Macfarlane, Mirvis, Pahlm, Rautaharju, and
  Wagner}]{doi:10.1161/CIRCULATIONAHA.106.180200}
Kligfield, P.; Gettes, L.~S.; Bailey, J.~J.; Childers, R.; Deal, B.~J.;
  Hancock, E.~W.; van Herpen, G.; Kors, J.~A.; Macfarlane, P.; Mirvis, D.~M.;
  Pahlm, O.; Rautaharju, P.; and Wagner, G.~S. 2007.
\newblock Recommendations for the Standardization and Interpretation of the
  Electrocardiogram.
\newblock \emph{Circulation}, 115(10): 1306--1324.

\bibitem[{Kwiatkowski et~al.(1992)Kwiatkowski, Phillips, Schmidt, and
  Shin}]{kwiatkowski1992testing}
Kwiatkowski, D.; Phillips, P.~C.; Schmidt, P.; and Shin, Y. 1992.
\newblock Testing the null hypothesis of stationarity against the alternative
  of a unit root: How sure are we that economic time series have a unit root?
\newblock \emph{Journal of econometrics}, 54(1-3): 159--178.

\bibitem[{Liu et~al.(2018)Liu, Liu, Zhao, Zhang, Wu, Xu, Liu, Ma, Wei, He
  et~al.}]{liu2018open}
Liu, F.; Liu, C.; Zhao, L.; Zhang, X.; Wu, X.; Xu, X.; Liu, Y.; Ma, C.; Wei,
  S.; He, Z.; et~al. 2018.
\newblock An open access database for evaluating the algorithms of
  electrocardiogram rhythm and morphology abnormality detection.
\newblock \emph{Journal of Medical Imaging and Health Informatics}, 8(7):
  1368--1373.

\bibitem[{Liu et~al.(2021)Liu, Zhang, Hou, Mian, Wang, Zhang, and
  Tang}]{Liu_2021}
Liu, X.; Zhang, F.; Hou, Z.; Mian, L.; Wang, Z.; Zhang, J.; and Tang, J. 2021.
\newblock Self-supervised Learning: Generative or Contrastive.
\newblock \emph{IEEE Transactions on Knowledge and Data Engineering}, 1–1.

\bibitem[{Ma et~al.(2019)Ma, Zheng, Li, and Cottrell}]{ma2019learning}
Ma, Q.; Zheng, J.; Li, S.; and Cottrell, G.~W. 2019.
\newblock Learning representations for time series clustering.
\newblock \emph{Advances in neural information processing systems}, 32:
  3781--3791.

\bibitem[{Mehari and Strodthoff(2021)}]{mehari2021self}
Mehari, T.; and Strodthoff, N. 2021.
\newblock Self-supervised representation learning from 12-lead ECG data.
\newblock \emph{arXiv preprint arXiv:2103.12676}.

\bibitem[{Mohsenvand, Izadi, and Maes(2020)}]{mohsenvand2020contrastive}
Mohsenvand, M.~N.; Izadi, M.~R.; and Maes, P. 2020.
\newblock Contrastive representation learning for electroencephalogram
  classification.
\newblock In \emph{Machine Learning for Health}, 238--253. PMLR.

\bibitem[{Oord, Li, and Vinyals(2018)}]{oord2018representation}
Oord, A. v.~d.; Li, Y.; and Vinyals, O. 2018.
\newblock Representation learning with contrastive predictive coding.
\newblock \emph{arXiv preprint arXiv:1807.03748}.

\bibitem[{Oppenheim, Willsky, and Nawab(1996)}]{signalsystem}
Oppenheim, A.~V.; Willsky, A.~S.; and Nawab, S.~H. 1996.
\newblock \emph{Signals \& Systems (2nd Ed.)}.
\newblock USA: Prentice-Hall, Inc.
\newblock ISBN 0138147574.

\bibitem[{Paszke et~al.(2019)Paszke, Gross, Massa, Lerer, Bradbury, Chanan,
  Killeen, Lin, Gimelshein, Antiga et~al.}]{paszke2019pytorch}
Paszke, A.; Gross, S.; Massa, F.; Lerer, A.; Bradbury, J.; Chanan, G.; Killeen,
  T.; Lin, Z.; Gimelshein, N.; Antiga, L.; et~al. 2019.
\newblock Pytorch: An imperative style, high-performance deep learning library.
\newblock \emph{Advances in neural information processing systems}, 32:
  8026--8037.

\bibitem[{Sarkar and Etemad(2020)}]{sarkar2020self}
Sarkar, P.; and Etemad, A. 2020.
\newblock Self-supervised ECG representation learning for emotion recognition.
\newblock \emph{IEEE Transactions on Affective Computing}.

\bibitem[{Spathis et~al.(2020)Spathis, Perez-Pozuelo, Brage, Wareham, and
  Mascolo}]{spathis2020learning}
Spathis, D.; Perez-Pozuelo, I.; Brage, S.; Wareham, N.~J.; and Mascolo, C.
  2020.
\newblock Learning Generalizable Physiological Representations from Large-scale
  Wearable Data.
\newblock \emph{arXiv preprint arXiv:2011.04601}.

\bibitem[{Tian, Krishnan, and Isola(2020)}]{tian2020contrastive}
Tian, Y.; Krishnan, D.; and Isola, P. 2020.
\newblock Contrastive Multiview Coding.
\newblock In Vedaldi, A.; Bischof, H.; Brox, T.; and Frahm, J.-M., eds.,
  \emph{Computer Vision -- ECCV 2020}, 776--794. Cham: Springer International
  Publishing.
\newblock ISBN 978-3-030-58621-8.

\bibitem[{Tonekaboni, Eytan, and Goldenberg(2021)}]{tonekaboni2021unsupervised}
Tonekaboni, S.; Eytan, D.; and Goldenberg, A. 2021.
\newblock Unsupervised Representation Learning for Time Series with Temporal
  Neighborhood Coding.
\newblock In \emph{International Conference on Learning Representations}.

\bibitem[{van~der Maaten and Hinton(2008)}]{tsne}
van~der Maaten, L.; and Hinton, G. 2008.
\newblock Visualizing Data using t-SNE.
\newblock \emph{Journal of Machine Learning Research}, 9(86): 2579--2605.

\bibitem[{Wagner et~al.(2020)Wagner, Strodthoff, Bousseljot, Kreiseler, Lunze,
  Samek, and Schaeffter}]{wagner2020ptb}
Wagner, P.; Strodthoff, N.; Bousseljot, R.-D.; Kreiseler, D.; Lunze, F.~I.;
  Samek, W.; and Schaeffter, T. 2020.
\newblock PTB-XL, a large publicly available electrocardiography dataset.
\newblock \emph{Scientific Data}, 7(1): 1--15.

\bibitem[{WHO.(2019)}]{who2019}
WHO. 2019.
\newblock Cardiovascular diseases (CVDs).

\bibitem[{Yue et~al.(2021)Yue, Wang, Duan, Yang, Huang, and
  Xu}]{yue2021learning}
Yue, Z.; Wang, Y.; Duan, J.; Yang, T.; Huang, C.; and Xu, B. 2021.
\newblock Learning Timestamp-Level Representations for Time Series with
  Hierarchical Contrastive Loss.
\newblock \emph{arXiv preprint arXiv:2106.10466}.

\bibitem[{Zheng et~al.(2020)Zheng, Zhang, Danioko, Yao, Guo, and
  Rakovski}]{zheng202012}
Zheng, J.; Zhang, J.; Danioko, S.; Yao, H.; Guo, H.; and Rakovski, C. 2020.
\newblock A 12-lead electrocardiogram database for arrhythmia research covering
  more than 10,000 patients.
\newblock \emph{Scientific data}, 7(1): 1--8.

\bibitem[{Zhu et~al.(2021)Zhu, Lan, Zhao, Guo, Kojodjojo, Xu, Liu, Liu, Wang,
  Sun, and Feng}]{Zhu_2021}
Zhu, Z.; Lan, X.; Zhao, T.; Guo, Y.; Kojodjojo, P.; Xu, Z.; Liu, Z.; Liu, S.;
  Wang, H.; Sun, X.; and Feng, M. 2021.
\newblock Identification of 27 abnormalities from multi-lead {ECG} signals: an
  ensembled {SE}{\_}{ResNet} framework with Sign Loss function.
\newblock \emph{Physiological Measurement}, 42(6): 065008.

\end{thebibliography}

\newpage
\appendix
\onecolumn

\setcounter{table}{0}
\setcounter{figure}{0}
\renewcommand{\thetable}{A\arabic{table}}
\renewcommand{\thefigure}{A\arabic{figure}}

\begin{subappendices}
\renewcommand{\thesubsection}{\Alph{section}.\arabic{subsection}}

\section{Appendix}

\subsection{Qualitative Analysis}
\begin{figure}[h!]
    \centering
    \includegraphics[width=0.8\textwidth]{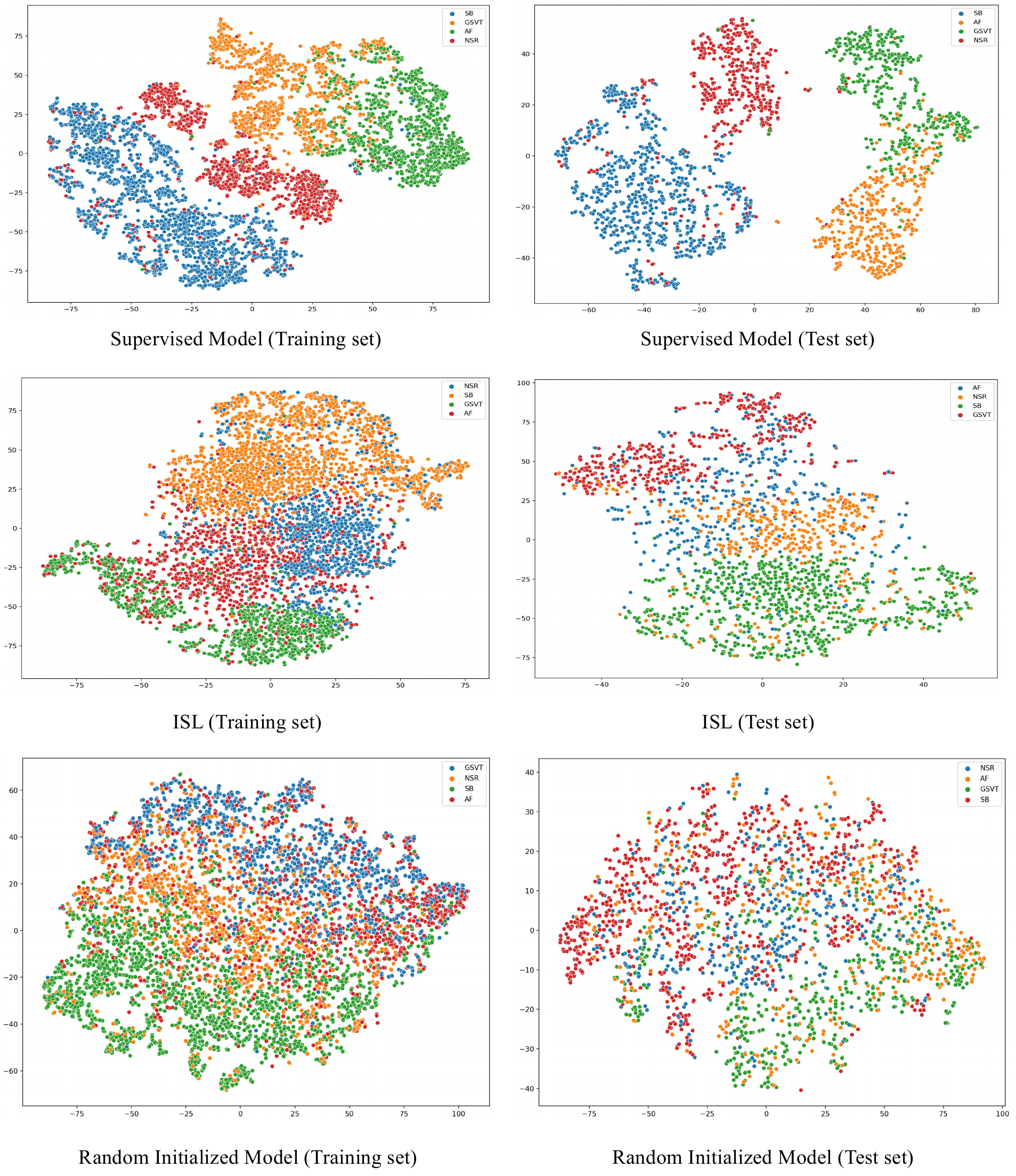}
    \caption{t-SNE visualization of representations for the Chapman dataset. Each data point in the plot is a 256-dimensional representation of a multivariate cardiac signal. The color represents different categories of cardiac arrhythmia in the dataset.} 
    \label{fig:tsne}
\end{figure}

Figure \ref{fig:tsne} shows the t-SNE visualization \cite{tsne} of representations for the Chapman dataset (four types of cardiac arrhythmia) across the supervised model, ISL, and randomly initialized ISL encoder in the linear evaluation scenario. We can see that ISL significantly improves representations' quality compared with the randomly initialized model. Furthermore, the clusterability of representations from ISL is close to the supervised model (the upper-bound of performance) as four clusters are clearly observed. This suggests that ISL extracts information-rich representations as expected.

\begin{figure}[h!]
    \centering
    \includegraphics[width=0.9\textwidth]{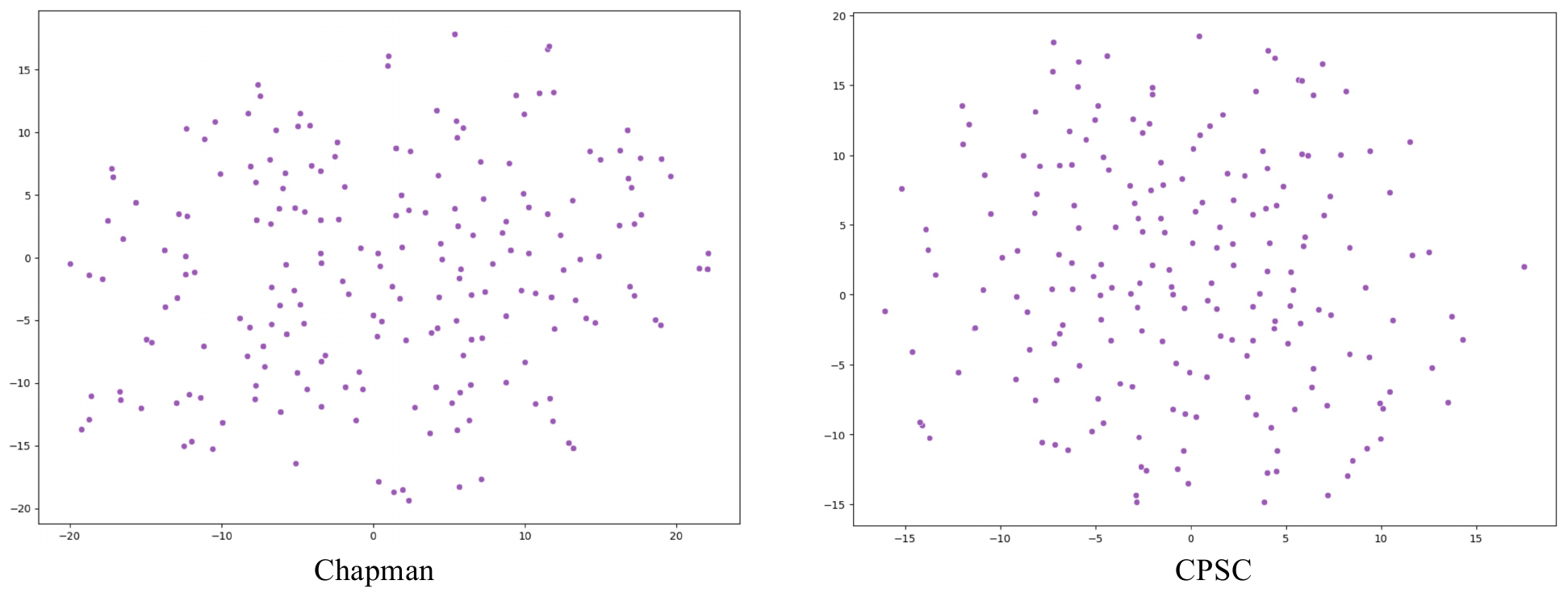}
    \caption{Distinctive representations for different patients. Each data point in the plot is a 256-dimensional representation of a unique patient.} 
    \label{fig:pat_tsne}
\end{figure}

We visualize the distinction of representations for different patients in Figure \ref{fig:pat_tsne}. We randomly sample 200 patients from the Chapman and CPSC datasets, respectively. Then we visualize their representations using the t-SNE plot. It can be observed that the representations are very scattered, and there are no clusters in the plot. This indicates ISL's capability of learning distinctive representations for different patients.

\newpage

\subsection{Data Pre-processing}

\begin{itemize}

\item \textbf{Chapman} \cite{zheng202012}.
Each ECG recording consists 5,000 data points ($i.e.$, 10 seconds duration with 500 Hz sampling rate). We divide each recording into ten equal length and non-overlapping frames, such that each frame consists of 500 data points. We remove the recordings that contain null values or more than 500 data points of constant values, as these recordings are likely to be abnormal data due to sampling errors. We follow the suggestion from \cite{zheng202012} and group the cardiac arrhythmias in the dataset into four classes: Atrial Fibrillation (AF), GSVT, Sudden Bradychardia (SB), Sinus Rhythm (SR). 

\item \textbf{CPSC} \cite{liu2018open}.
The duration of ECG recordings varies from 6 seconds to 60 seconds with 500 Hz sampling rate. We remove the recordings whose duration is less than 10 seconds. For recordings longer than 10 seconds, we truncate them to multiple segments that each segment is 10 seconds. Then we apply the same processing and exclusion criterion to each segment as we applied to the Chapman dataset. There is a total of nine types of cardiac arrhythmias in the dataset: Atrial Fibrillation, Normal Sinus Rhythm, Premature Atrial Contractions, Premature Ventricular Contractions, Left Bundle Branch Block, Right Bundle Branch Block, ST-segment depression, and ST-segment elevated. Multiple labels are assigned to each ECG recording.

\item \textbf{PTB-XL} \cite{wagner2020ptb}. 
Each ECG recording consists 5,000 data points ($i.e.$, 10 seconds duration with 500 Hz sampling rate). Therefore, we apply the same processing and exclusion criterion to each recording as we applied to the Chapman dataset. There are 71 cardiac arrhythmias in the dataset, and multiple labels are assigned to each ECG recording.

\end{itemize}

\subsection{Pre-training, Fine-tuning and Evaluation setup}
We split each dataset into 60\%, 20\%, 20\% in subject-wise for training, validation, and testing. In all experiments, we use a logistic regression model as the classifier.

\begin{itemize}

\item \textbf{Linear evaluation}. We pre-train ISL on the training set of a source dataset first. Then we train the classifier on the training set to identify cardiac arrhythmias in a supervised manner, using features extracted from the frozen ISL encoder. We report AUROC on the test set of the source dataset.

\item \textbf{Transferability evaluation}. We pre-train ISL on the entire source dataset. After that, we fine-tune the ISL encoder and the classifier on the training set of the other two target datasets to identify cardiac arrhythmias in a supervised manner. We report AUROC on the test set of target datasets.

\item \textbf{Semi-supervised learning experiments}. We pre-train ISL on the entire source dataset. Later on, we fine-tune the ISL encoder and the classifier on the training set of another target dataset with different percentages of labeled training data to identify cardiac arrhythmias. We report AUROC on the test set of the target dataset.

\end{itemize}

\newpage

\subsection{Data Augmentations}
\begin{figure}[h]
    \centering
    \includegraphics[width=0.8\textwidth]{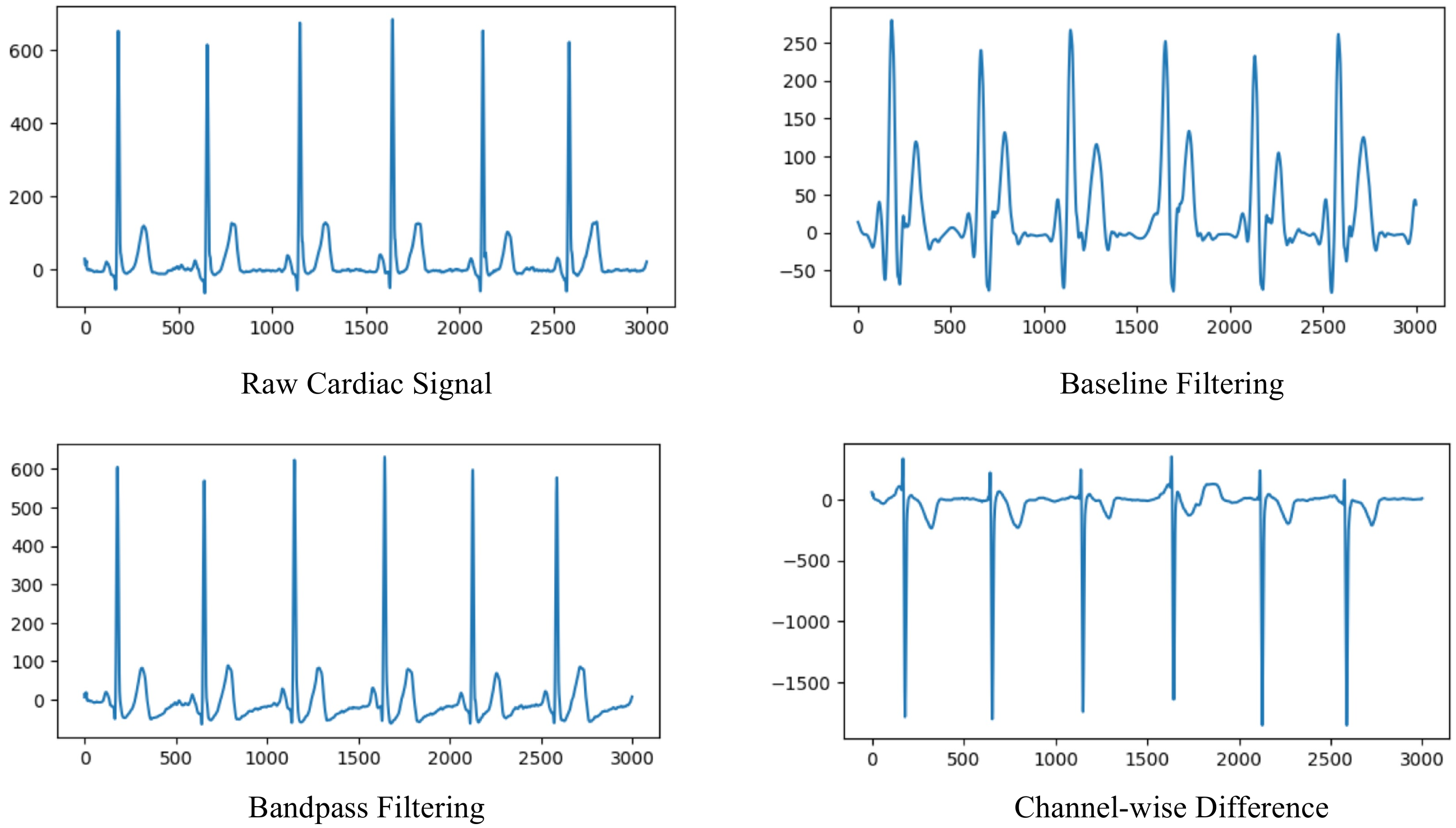}
    \caption{Example of data augmentations.}
    \label{fig:aug}
\end{figure}

For each input multivariate cardiac signal $\boldsymbol{X} \in R^{H \times L}$, we apply two data augmentations randomly sampled from following augmentations. Some examples of data augmentations shown in Figure \ref{fig:aug}.

\begin{itemize}
    \item \textbf{Baseline filtering}
        \begin{itemize}
            \item{Step 1.} We apply fifth-order Daubechies (db5) wavelet transform to $h$-th channel $\boldsymbol{X}_h$ of the input signal $\boldsymbol{X}$. We can obtain a list of coefficients arrays $\{A_5, D_5, D_4, D_3, D_2, D_1\}$ from this transformation. $A_5$ is the approximation coefficients array (capture low-frequency information) from the fifth level of decomposition, and $D_5$ to $D_1$ are detail coefficients arrays (capture high-frequency information) of each decomposition level. 
            \item{Step 2.} We set detail coefficients arrays to zero, now the coefficients arrays' list is $\{A_5, 0, 0, 0, 0, 0\}$. Then we reconstruct the signal $\boldsymbol{\widehat{X}}_h$ from the list using the inverse discrete wavelet (db5) transform.
            \item{Step 3.} Repeat Step 1 and Step 2 for each channel in $\boldsymbol{X}$ to get the transformed signal $\boldsymbol{\widehat{X}} \in R^{H \times L}$.
        \end{itemize}
    
    \item \textbf{Bandpass filtering}
        \begin{itemize}
            \item{Step 1.} We apply Finite Impulse Response (FIR) bandpass filter \cite{signalsystem} to $h$-th channel $\boldsymbol{X}_h$ of the input signal $\boldsymbol{X}$. This decompose the signal into low (0.001-0.5 Hz), middle (0.5-50 Hz), and high-frequency ($>$50 Hz) signals $\boldsymbol{\widehat{X}}_h$ = \{$\boldsymbol{\widehat{X}}_h^{low}$, $\boldsymbol{\widehat{X}}_h^{middle}$, $\boldsymbol{\widehat{X}}_h^{high}$\}.
        
            \item{Step 2.} Repeat Step 1 for each channel in $\boldsymbol{X}$ to get the transformed signal $\boldsymbol{\widehat{X}} \in R^{3H \times L}$, and we select the middle-frequency band only $\boldsymbol{\widehat{X}} = \{\boldsymbol{\widehat{X}}_1^{middle}, \boldsymbol{\widehat{X}}_2^{middle}, \hdots, \boldsymbol{\widehat{X}}_H^{middle}\}$.
        \end{itemize}
    
    \item \textbf{Channel-wise difference}
        \begin{itemize}
            \item Step 1. We subtract $h$-th channel $\boldsymbol{X}_h$ from it's adjacent channel $\boldsymbol{X}_{h+1}$ to get a new channel  $\boldsymbol{\widehat{X}}_h$. Note that the last channel $\boldsymbol{\widehat{X}}_H$ = $\boldsymbol{X}_H$ - $\boldsymbol{X}_1$.
            \item Step 2. Repeat Step 1 for each channel in $\boldsymbol{X}$ to get the transformed signal $\boldsymbol{\widehat{X}} \in R^{H \times L}$. 
        \end{itemize}

    \item \textbf{Amplitude Scaling}
        \begin{itemize}
            \item Step 1. Random sample a scaling ratio $\alpha$ from 0.5 to 2.
            \item Step 2. Multiply each channel in $\boldsymbol{X}$ with $\alpha$ to get the transformed signal $\boldsymbol{\widehat{X}} \in R^{H \times L}$. 
        \end{itemize}
    
    \item \textbf{Amplitude Reversing}
        \begin{itemize}
            \item Step 1. Multiply each channel in $\boldsymbol{X}$ with $-1$ to get the transformed signal $\boldsymbol{\widehat{X}} \in R^{H \times L}$. 
    \end{itemize}
\end{itemize}

\end{subappendices}

\end{document}